\documentclass{article}

\usepackage[utf8]{inputenc}
\usepackage[T1]{fontenc}

\usepackage[final]{neurips_2023}
\usepackage{natbib}
\setcitestyle{numbers,square}

\usepackage{hyperref}

\usepackage{url}
\usepackage{booktabs}
\usepackage{amsfonts}
\usepackage{nicefrac}
\usepackage{microtype}
\usepackage{xcolor}
\usepackage{multirow}
\usepackage{graphicx}
\usepackage{CJKutf8}

\title{MiChao-HuaFen 1.0: A Specialized Pre-trained Corpus Dataset for Domain-specific Large Models}

\author{%
  Yidong Liu \\
  Shanghai Midu Technology Co., Ltd \\
  \texttt{yidong@midu.com} \\
  \And
  FuKai Shang \\
  Shanghai AI Laboratory \\
  \texttt{shangfukai@pjlab.org.cn} \\
  \And
  Fang Wang \\
  Shanghai Midu Technology Co., Ltd \\
  \texttt{wangfang@midu.com} \\
  \And
  Rui Xu \\
  Shanghai AI Laboratory \\
  \texttt{xurui@pjlab.org.cn} \\
  \And
  Jun Wang \\
  Shanghai Midu Technology Co., Ltd \\
  \texttt{wangjun@midu.com} \\
  \And
  Wei Li \\
  Shanghai AI Laboratory \\
  \texttt{liwei@pjlab.org.cn} \\
  \And
  Yao Li \\
  Shanghai Midu Technology Co., Ltd \\
  \texttt{liyao@midu.com} \\
  \And
  Conghui He \\
  Shanghai AI Laboratory \\
  \texttt{heconghui@pjlab.org.cn} \\
  \And
}

\begin{document}
\maketitle

\begin{abstract}
With the advancement of deep learning technologies, general-purpose large models such as GPT-4 have demonstrated exceptional capabilities across various domains. Nevertheless, there remains a demand for high-quality, domain-specific outputs in areas like healthcare, law, and finance. This paper first evaluates the existing large models for specialized domains and discusses their limitations. To cater to the specific needs of certain domains, we introduce the ``MiChao-HuaFen 1.0'' pre-trained corpus dataset, tailored for the news and governmental sectors. The dataset, sourced from publicly available internet data from 2022, underwent multiple rounds of cleansing and processing to ensure high quality and reliable origins, with provisions for consistent and stable updates. This dataset not only supports the pre-training of large models for Chinese vertical domains but also aids in propelling deep learning research and applications in related fields.
\end{abstract}

\section{Introduction}
In the realm of general-purpose large models, models like GPT-4\cite{openai2023gpt4}  have showcased formidable comprehensive capabilities, ranging from general knowledge Q\&A, content creation, coding, to reasoning, often matching or even surpassing human abilities. However, these general models still exhibit deficiencies in domain-specific knowledge, especially in areas like healthcare, law, and finance. This necessitates large models tailored for specific domains to achieve outputs that align with domain expertise, especially in the Chinese context. Currently, there are several studies on large models for specific domains, for instance, in healthcare: DoctorGLM\cite{doctorglm2023}, Huatuo-Llama-Med-Chinese\cite{calla2023}; in law: LaWGPT\cite{lawgpt2023}, ChatLaw\cite{chatlaw2023}, Lawyer LLaMA\cite{lawyerllama2023}; and in finance: FinGPT\cite{fingpt2023}, Cornucopia-LLaMA-Fin-Chinese\cite{cornucopia2023}. Most of these models are fine-tuned from open-source base models like ChatGLM and LLama. Studies have shown\cite{wei2023} that without incorporating relevant corpora during the pre-training phase and relying solely on fine-tuning, optimal model performance cannot be achieved. Hence, this research introduces the "MiChao-HuaFen 1.0" pre-trained corpus dataset for news and governmental domain models, aiming to better support the corpus requirements during the pre-training phase of Chinese vertical domain models. While there are existing open-source pre-trained corpus datasets, such as "WanJuan 1.0"\cite{wanjuan2023} by OpenDataLab\cite{opendatalab2022}, "MiChao-HuaFen" focuses on collecting data from news and governmental sources, curated from publicly accessible websites' historical data from 2022, ensuring reliable origins, high data quality, and consistent updates.

\section{Dataset Statistics}
The "MiChao-HuaFen 1.0" corpus is sourced from publicly accessible Chinese internet data, primarily from news and governmental domains. Through keyword filtering, image extraction, rule-based filtering, and format conversion, a high-quality text model corpus has been established. The final cleaned data consists of over 70 million entries, including over 1 million image links. Sample data is as follows: 

\begin{figure}
\centering
\includegraphics[width=1\linewidth]{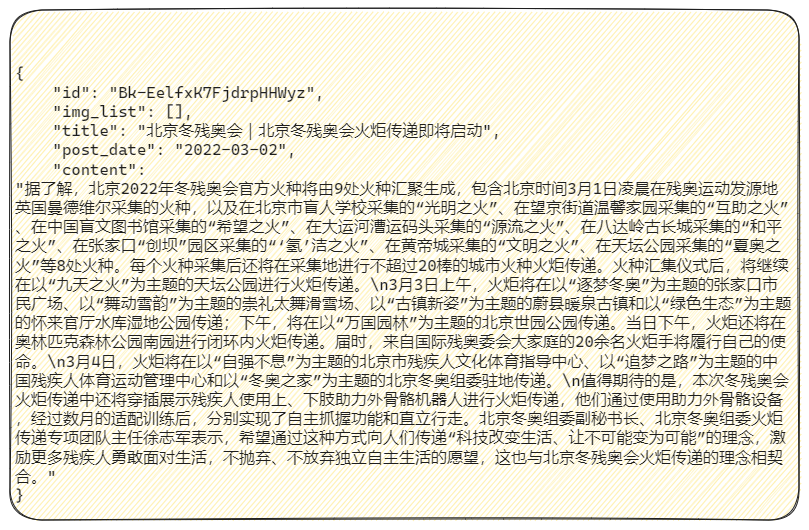}
\caption{\label{fig:sample}Data Sample.}
\end{figure}

\begin{itemize}
\item id: Unique document ID (String type).
\item img\_list: List of image URLs within the document (Array type).
\item title: Document title (String type), in plain text or Markdown format.
\item post\_date: Document publication date (String type).
\item content: Document content (String type), in plain text or Markdown format.

\end{itemize}

\section{Methods}
The data processing methods for "MiChao-HuaFen 1.0" largely adhere to existing pre-trained corpus cleaning methodologies, primarily following these principles:
\begin{itemize}
\item Source compliance: Initial data source screening ensures licensing and compliance, with language filters to guarantee Chinese content.Corpus diversity: Aim to select diverse types and sources of data.
\item Sensitive information removal: Keyword filtering and model classification are employed to filter out sensitive data sources and remove PII.
\item Ensuring corpus quality: Multiple data processing techniques and rounds of review ensure corpus purity and quality, removing unsuitable training data based on our model training experience. 
\end{itemize}
Based on these principles, the corpus underwent the following processing: 

\begin{figure}
\centering
\includegraphics[width=1\linewidth]{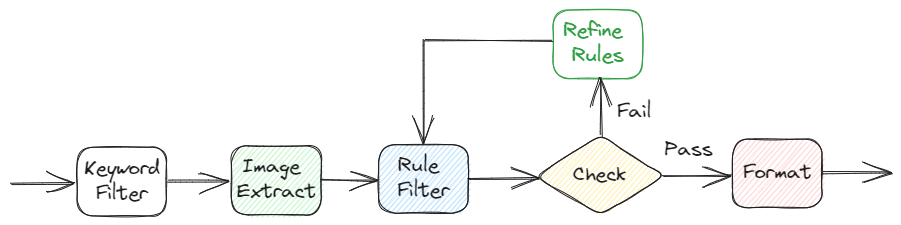}
\caption{\label{fig:MiChao_processing}Data processing pipeline.}
\end{figure}

\begin{enumerate}
\item Keyword Filtering: As a primary step in web content processing, keyword filtering is crucial for ensuring content safety and accuracy. By building a keyword library, we can swiftly identify and remove content with sensitive or inappropriate terms, enhancing data quality and ensuring user safety.
\item 	Image Extraction: Images, as information carriers, often convey more than plain text. Thus, we extracted image links from the corpus. Using xpath technology, we can efficiently extract image links from complex web structures. All extracted image links are stored in an array field for further processing and application.
\item Rule-based Filtering: This key step refines web content to ensure corpus conciseness and relevance. We employed numerous rule strategies, such as using xpath to remove all HTML tags like <script> and <style>, and filtering out content shorter than 200 characters to ensure depth and information richness.
\item Quality Inspection: To further ensure data quality, we combined manual and automated model quality checks. Some data undergo manual sampling, while models scan the entire dataset to ensure accuracy and completeness.
\item Rule Refinement: During actual processing, there are always exceptions or special cases. For data that doesn't meet requirements, we further refine processing rules, adding new rules to our rule library. This way, the processing system continually learns and evolves, adapting to more scenarios and needs.
\item Formatting: The final step is to format the processed data, ensuring it adheres to standard Markdown format. We also retain previously extracted image links, providing content with both text and rich image information.
\end{enumerate}

\section{Conclusion}
As large models become more prevalent, providing them with more specialized and targeted pre-training data has become a pivotal research direction. The release of the "MiChao-HuaFen 1.0" pre-trained corpus dataset aims to meet this demand, especially in the news and governmental verticals. The primary audience for this dataset includes, but is not limited to:

\begin{itemize}
\item AI researchers and scholars: For those researching Chinese domain model pre-training, this dataset offers more professional and high-quality pre-training corpora.
\item Enterprises and institutions: Especially news agencies and government departments can utilize this dataset for their model pre-training, achieving more precise and compliant model outputs. 
\end{itemize}

The ``MiChao-HuaFen 1.0'' pre-trained corpus dataset can be accessed and downloaded at \url{https://opendatalab.org.cn/OpenDataLab/WanJuan1_dot_0}.

To ensure data compliance and privacy, we rigorously screened and reviewed all content before release, ensuring no sensitive information is included. However, it's crucial to note that any organization or individual using the dataset must adhere to the respective usage agreement and cite our related publications when conducting research or applications.

\bibliographystyle{plain}
\bibliography{references}

\end{document}